\documentclass[11pt,a4paper]{article}
\usepackage[hyperref]{emnlp-ijcnlp-2019}
\usepackage{times}
\usepackage{latexsym}
\usepackage{bm}
\usepackage{amsfonts}
\usepackage{amsmath}
\usepackage{xspace}
\usepackage{url}
\usepackage{multirow}
\usepackage{graphicx}

\usepackage{enumitem}

\usepackage{booktabs}

\aclfinalcopy 

\newcommand\OurModel{NumNet\xspace}

\title{NumNet: Machine Reading Comprehension with Numerical Reasoning}

\author{Qiu Ran$^1$\thanks{\ \ indicates equal contribution}, Yankai Lin$^1$\footnotemark[1], Peng Li$^1$, Jie Zhou$^1$, Zhiyuan Liu$^2$ \\
  $^1$Pattern Recognition Center, WeChat AI, Tencent Inc, China \\
  $^{2}$Department of Computer Science and Technology, Tsinghua University, Beijing, China\\
Institute for Artificial Intelligence, Tsinghua University, Beijing, China\\
State Key Lab on Intelligent Technology and Systems, Tsinghua University, Beijing, China \\
  \texttt{\{soulcaptran,yankailin,patrickpli,withtomzhou\}@tencent.com}\\
  \texttt{liuzy@tsinghua.edu.cn}
}
\date{}

\begin{document}
\maketitle
\begin{abstract}
Numerical reasoning, such as addition, subtraction, sorting and counting is a critical skill in human's reading comprehension, which has not been well considered in existing machine reading comprehension (MRC) systems. To address this issue, we propose a numerical MRC model named as \OurModel, which utilizes a numerically-aware graph neural network to consider the comparing information and  performs numerical reasoning over numbers in the question and passage. Our system achieves an EM-score of $64.56\%$ on the DROP dataset, outperforming all existing machine reading comprehension models by considering the numerical relations among numbers.

\end{abstract}

\section{Introduction}
\label{sec:intro}

Machine reading comprehension (MRC) aims to infer the answer to a question given the document. In recent years, researchers have proposed lots of MRC  models~\citep{chen-bolton-manning:2016:P16-1,dhingra-EtAl:2017:Long2,cui-EtAl:2017:Long,seo2017bidirectional} and these models have achieved remarkable results in various public benchmarks such as SQuAD~\citep{rajpurkar2016squad} and  RACE~\citep{lai2017race}. The success of these models is due to two reasons: (1) Multi-layer architectures which allow these models to read the document and the question iteratively for reasoning; (2) Attention mechanisms which would enable these models to focus on the part related to the question in the document.

However, most of existing MRC models are still weak in numerical  reasoning such as addition, subtraction, sorting and counting~\citep{dua2019drop}, which are naturally required when reading financial news, scientific articles, etc. \citet{dua2019drop} proposed a numerically-aware QANet (NAQANet) model, which divides the answer generation for numerical MRC into three types: (1) extracting spans; (2) counting; (3) addition or subtraction over numbers. NAQANet makes a pioneering attempt to answer numerical questions but still does not explicitly consider numerical reasoning.

To tackle this problem, we introduce a novel model \OurModel that integrates numerical reasoning into existing MRC models. A key problem to answer questions requiring numerical reasoning is how to perform numerical comparison in MRC systems, which is crucial for two common types of questions:

(1) \textbf{Numerical Comparison}: The answers of the questions can be directly obtained via performing numerical comparison, such as sorting and comparison, in the documents. For example, in Table~\ref{tab:example}, for the first question, if the MRC system knows the fact that ``$49>47>36>31>22$'', it could easily extract that the second longest field goal is 47-yard. 

(2) \textbf{Numerical Condition}: The answers of the questions cannot be directly obtained through simple numerical comparison in the documents, but often require numerical comparison for understanding the text. For example, for the second question in Table~\ref{tab:example}, an MRC system needs to know which age group made up more than 7\% of the population to count the group number.  

\begin{table*}[ht]
\small
\center
\begin{tabular}{p{0.15\textwidth}p{0.65\textwidth}c}
\toprule
\textbf{Question} & \textbf{Passage} & \textbf{Answer}\\ 
\midrule
What is the second longest field goal made? & ... The Seahawks immediately trailed on a scoring rally by the Raiders with kicker \textit{\color{red} Sebastian Janikowski nailing a 31-yard field goal} ... Then in the third quarter \textit{\color{red} Janikowski made a 36-yard field goal}. Then \textit{\color{red} he made a 22-yard field goal} in the fourth quarter to put the Raiders up 16-0 ... The Seahawks would make their only score of the game with kicker \textit{\color{red} Olindo Mare hitting a 47-yard field goal}. However, they continued to trail as \textit{\color{red} Janikowski made a 49-yard field goal}, followed by RB Michael Bush making a 4-yard TD run. & 47-yard\\
\midrule
How many age groups made up more than 7\% of the population? &Of Saratoga Countys population in 2010, \textit{\color{blue} 6.3\%} were between ages of 5 and 9 years, \textit{\color{blue}6.7\%} between 10 and 14 years, 6.5\% between 15 and 19 years, \textit{\color{blue}5.5\%} between 20 and 24 years, \textit{\color{blue}5.5\%} between 25 and 29 years, \textit{\color{blue}5.8\%} between 30 and 34 years, \textit{\color{blue}6.6\%} between 35 and 39 years, \textit{\color{blue}7.9\%} between 40 and 44 years, \textit{\color{blue}8.5\%} between 45 and 49 years, \textit{\color{blue}8.0\%} between 50 and 54 years, \textit{\color{blue}7.0\%} between 55 and 59 years, \textit{\color{blue}6.4\%} between 60 and 64 years, and \textit{\color{blue}13.7\%} of age 65 years and over ... & 5\\
\bottomrule

\end{tabular}
\caption{Example questions from the DROP dataset which require numerical comparison. We highlight the relevant parts in the passage to infer the answer.}
\label{tab:example}
\end{table*}

Hence, our \OurModel model considers numerical comparing information among numbers when answering numerical questions. As shown in Figure~\ref{fig:main}, \OurModel first encodes both the question and passages through an encoding module consisting of convolution layers, self-attention layers and feed-forward layers as well as a passage-question attention layer. After that, we feed the question and passage representations into a numerically-aware graph neural network (NumGNN) to further integrate the comparison information among numbers into their representations. Finally, we utilize the numerically-aware representation of passages to infer the answer to the question.

The experimental results on a public numerical MRC dataset DROP~\citep{dua2019drop} show that our \OurModel model achieves significant and consistent improvement as compared to all baseline methods by explicitly performing numerical reasoning over numbers in the question and passage. In particular, we show that our model could effectively deal with questions requiring sorting with multi-layer NumGNN. The source code of our paper is available at \url{https://github.com/ranqiu92/NumNet}.
\section{Related Work}
\label{sec:related_work}
\subsection{Machine Reading Comprehension}    
Machine reading comprehension (MRC) has become an important research area in NLP. In recent years, researchers have published a large number of annotated MRC datasets such as CNN/Daily Mail~\citep{hermann2015teaching}, SQuAD~\citep{rajpurkar2016squad}, RACE~\citep{lai2017race}, TriviaQA~\citep{joshi-EtAl:2017:Long} and so on. With the blooming of available large-scale MRC datasets,  a great number of neural network-based MRC models have been proposed to answer questions for a given document including Attentive Reader~\citep{kadlec2016text}, BiDAF~\citep{seo2017bidirectional}, Interactive AoA Reader~\citep{cui-EtAl:2017:Long}, Gated Attention Reader~\citep{dhingra-EtAl:2017:Long2}, R-Net~\citep{wang2017gated}, DCN~\citep{xiong2017dynamic}, QANet~\citep{yu2018qanet}, and achieve promising results in most existing public MRC datasets.

Despite the success of neural network-based MRC models, researchers began to analyze the data and rethink to what extent we have solved the problem of MRC. Some works~\citep{chen-bolton-manning:2016:P16-1,sugawara2018makes,kaushik2018much} classify the reasoning skills required to answer the questions into the following types: (1) Exact matching/Paraphrasing; (2) Summary; (3) Logic reasoning; (4) Utilizing external knowledge; (5) Numerical reasoning. They found that most existing MRC models are focusing on dealing with the first three types of questions. However, all these models suffer from problems when answering the questions requiring numerical reasoning. To the best of our knowledge, our work is the first one that \emph{explicitly} incorporates numerical reasoning into the MRC system.
The most relevant work to ours is NAQANet~\citep{dua2019drop}, which adapts the output layer of QANet~\citep{yu2018qanet} to support predicting answers based on counting and addition/subtraction over numbers. However, it does not consider numerical reasoning explicitly during encoding or inference.
 
\begin{figure*}
    \centering
    \includegraphics[width=0.98\textwidth]{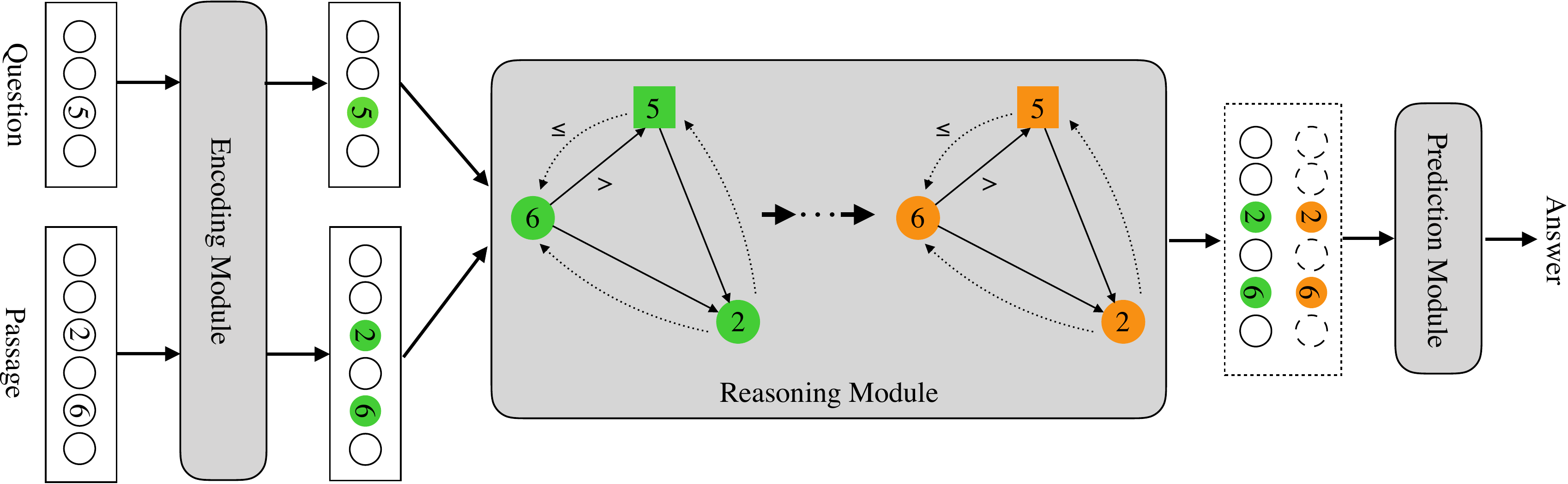}
    \caption{{\bf The framework of our \OurModel model.} Our model consists of an encoding module, a reasoning module and a prediction module. The numerical relations between numbers are encoded with the topology of the graph. For example, the edge pointing from ``6'' to ``5'' denotes ``6'' is greater than ``5''. And the reasoning module leverages a numerically-aware graph neural network to perform numerical reasoning on the graph. As numerical comparison is modeled explicitly in our model, it is more effective for answering questions requiring numerical  reasoning such as addition, counting, or sorting over numbers.}
    \label{fig:main}
\end{figure*}
    
\subsection{Arithmetic Word Problem Solving} 
Recently, understanding and solving arithmetic word problems (AWP) has attracted the growing interest of NLP researchers. \citet{hosseini2014learning} proposed a simple method to address arithmetic word problems, but mostly focusing on subsets of problems which only require addition and subtraction. After that, \citet{roy2015solving} proposed an algorithmic approach which could handle arithmetic word problems with multiple steps and operations. \citet{koncel2015parsing} further formalized the AWP problem as that of generating and scoring equation trees via integer linear programming. \citet{wang-etal-2017-deep-neural} and \citet{wang2017AWP} proposed sequence to sequence solvers for the AWP problems, which are capable of generating unseen expressions and do not rely on sophisticated manual features.
\citet{wang2018mathdqn} leveraged deep Q-network to solve the AWP problems, achieving a good balance between effectiveness and efficiency. However, all the existing AWP systems are only trained and validated on small benchmark datasets. \citet{huang2016well} found that the performance of these AWP systems sharply degrades on larger datasets. 
Moreover, from the perspective of NLP, MRC problems are more challenging than AWP since the passages in MRC are mostly real-world texts which require more complex skills to be understood. Above all, it is nontrivial to adapt most existing AWP models to the MRC scenario. Therefore, we focus on enhancing MRC models with numerical reasoning abilities in this work. 

\section{Methodology} 
\label{sec:method}
In this section, we will introduce the framework of our model \OurModel and provide the details of the proposed numerically-aware graph neural network (NumGNN) for numerical  reasoning.

\subsection{Framework}
\label{sec:framework}
An overview of our model \OurModel is shown in Figure~\ref{fig:main}. We compose our model with encoding module, reasoning module and prediction module. Our major contribution is the reasoning module, which leverages a NumGNN between the encoding module and prediction module to explicitly consider the numerical comparison information and perform numerical reasoning. As NAQANet has been shown effective for handling numerical MRC problem~\cite{dua2019drop}, we leverage it as our base model and mainly focus on the design and integration of the NumGNN in this work.

\paragraph{Encoding Module} Without loss of generality, we use the encoding components of QANet and NAQANet to encode the question and passage into vector-space representations. Formally, the question $Q$ and passage $P$ are first encoded as:
\begin{eqnarray}
    \bm{Q} &=& \texttt{QANet-Emb-Enc}(Q),\\
    \bm{P} &=& \texttt{QANet-Emb-Enc}(P),
\end{eqnarray}
and then the passage-aware question representation and the question-aware passage representation are computed as:
\begin{eqnarray}
    \bar{\bm{Q}} &=& \texttt{QANet-Att}(\bm{P},\bm{Q}),\\
    \bar{\bm{P}} &=& \texttt{QANet-Att}(\bm{Q},\bm{P}),
\end{eqnarray}
where $\texttt{QANet-Emb-Enc}(\cdot)$ and  $\texttt{QANet-Att}(\cdot)$ denote the ``stacked embedding encoder layer'' and ``context-query attention layer'' of QANet respectively. The former consists of convolution, self-attention and feed-forward layers. The latter is a passage-question attention layer. $\bar{\bm{Q}}$ and $\bar{\bm{P}}$ are used by the following components.

\paragraph{Reasoning Module} First we build a heterogeneous directed graph $\mathcal{G}=(\bm{V};\bm{E})$, whose nodes ($\bm{V}$) are corresponding to the numbers in the question and passage, and edges ($\bm{E}$) are used to encode numerical relationships among the numbers. The details will be explained in Sec.~\ref{sec:graph}. 

Then we perform reasoning on the graph based on a graph neural network, which can be formally denoted as:
\begin{eqnarray}
    \bm{M}^Q &=&\texttt{QANet-Mod-Enc}(\bm{W}^M\bar{\bm{Q}}),\\
    \bm{M}^P &=&\texttt{QANet-Mod-Enc}(\bm{W}^M\bar{\bm{P}}),\\
    \bm{U}&=&\texttt{Reasoning}(\mathcal{G}; \bm{M}^Q, \bm{M}^P),
    \label{eqn:reasoing}
\end{eqnarray}
where $\bm{W}^M$ is a shared weight matrix, $\bm{U}$ is the representations of the nodes corresponding to the numbers, $\texttt{QANet-Mod-Enc}(\cdot)$ is the ``model encoder layer'' defined in QANet which is similar to $\texttt{QANet-Emb-Enc}(\cdot)$, and the definition of $\texttt{Reasoning}(\cdot)$ will be given in Sec.~\ref{sec:reasoning}.

Finally, as $\bm{U}$ only contains the representations of numbers, to tackle span-style answers containing non-numerical words, we concatenate $\bm{U}$ with $\bm{M}^P$ to produce numerically-aware passage representation $\bm{M}_0$. Formally,
\begin{eqnarray}
    \bm{M}^{\text{num}}[i]&=&
        \left\{
            \begin{array}{ll}
                \bm{U}[I(i)] & \text{if $w^p_i$ is a number}\\
                \bm{0}
            \end{array}
        \right., \nonumber\\
    \bm{M}_0'&=&\bm{W}_0[\bm{M}^P;\bm{M}^{\text{num}}] + \bm{b}_0,\\
    \bm{M}_0&=&\texttt{QANet-Mod-Enc}(\bm{M}_0'),
\end{eqnarray}
where $[\cdot;\cdot]$ denotes matrix concatenation, $\bm{W}[k]$ denotes the $k$-th column of a matrix $\bm{W}$, $\bm{0}$ is a zero vector, $I(i)$ denotes the node index corresponding to the passage word $w_i^p$ which is a number, $\bm{W}_0$ is a weight matrix, and $\bm{b}_0$ is a bias vector.

\paragraph{Prediction Module} Following NAQANet~\citep{dua2019drop}, we divide the answers into four types and use a unique output layer to calculate the conditional answer probability $\Pr(\text{answer}|\text{type})$ for each type :
\begin{itemize}
    \item \emph{Passage span}: The answer is a span of the passage, and the answer probability is defined as the product of the probabilities of the start and end positions.
    
    \item \emph{Question span}: The answer is a span of the question, and the answer probability is also defined as the product of the probabilities of the start and end positions.
    
    \item \emph{Count}: The answer is obtained by counting, and it is treated as a multi-class classification problem over ten numbers ($0$-$9$), which covers most of the \emph{Count} type answers in the DROP dataset.
    
    \item \emph{Arithmetic expression}: The answer is the result of an arithmetic expression. The expression is obtained in three steps: (1) extract all numbers from the passage; (2) assign a sign (plus, minus or zero) for each number; (3) sum the signed numbers \footnote{As few samples require multiplication/division expression in the DROP dataset, we simply adapt the module proposed~\citep{dua2019drop} and leave multiplication/division expression handling as future work.}.
\end{itemize}

Meanwhile, an extra output layer is also used to predict the probability $\Pr(\text{type})$ of each answer type. At training time, the final answer probability is defined as the joint probability over all feasible answer types, i.e.,  $\sum_{\text{type}}\Pr(\text{type})\Pr(\text{answer}|\text{type})$. Here, the answer type annotation is not required and the probability $\Pr(\text{type})$ is learnt by the model. At test time, the model first selects the most probable answer type greedily and then predicts the best answer accordingly. 

Without loss of generality, we leverage the definition of the five output layers in~\citep{dua2019drop}, with $\bm{M_0}$ and $\bm{Q}$ as inputs. Please refer to the paper for more details due to space limitation.

\paragraph{Comparison with NAQANet}  The major difference between our model and NAQANet is that NAQANet does not have the reasoning module, i.e., $\bm{M}_0$ is simply set as $\bm{M}^P$. As a result, numbers are treated as common words in NAQANet except in the prediction module, thus NAQANet may struggle to learn the numerical relationships between numbers, and potentially cannot well generalize to unseen numbers.
However, as discussed in Sec.~\ref{sec:intro}, the numerical comparison is essential for answering questions requiring numerical reasoning. In our model, the numerical relationships are explicitly represented with the topology of the graph and a NumGNN is used to perform numerical reasoning. Therefore, our \OurModel model can handle questions requiring numerical reasoning more effectively, which is verified by the experiments in Sec.~\ref{sec:exp}.

\subsection{Numerically-aware Graph Construction}
\label{sec:graph}

We regard all numbers from the question and passage as nodes in the graph for reasoning \footnote{As a number in the question may serve as a critical comparison condition (refer to the second example in Table~\ref{tab:example}), we also add nodes for them in the graph.} . The set of nodes corresponding to the numbers occurring in question and passage are denoted as $\bm{V}^Q$ and $\bm{V}^P$ respectively. And we denote all the nodes as $\bm{V}=\bm{V}^Q\cup\bm{V}^P$, and the number corresponding to a node $v\in\bm{V}$ as $n(v)$.

Two sets of edges are considered in this work:
\begin{itemize}
    \item {\bf Greater Relation Edge ($\overrightarrow{\bm{E}}$}): For two nodes $v_i, v_j\in \bm{V}$, a directed edge $\overrightarrow{e}_{ij}=(v_i, v_j)$ pointing from $v_i$ to $v_j$ will be added to the graph if $n(v_i)>n(v_j)$, which is denoted as solid arrow in Figure~\ref{fig:main}.
    \item {\bf Lower or Equal Relation Edge ($\overleftarrow{\bm{E}}$)}: For two nodes $v_i, v_j\in \bm{V}$, a directed edge $\overleftarrow{e}_{ij}=(v_j, v_i)$ will be added to the graph if $n(v_i)\leq n(v_j)$, which is denoted as dashed arrow in Figure~\ref{fig:main}.
\end{itemize}
Theoretically, $\overrightarrow{\bm{E}}$ and $\overleftarrow{\bm{E}}$ are complement to each other . However, as a number may occur several times and represent different facts in a document, we add a distinct node for each occurrence in the graph to prevent potential ambiguity. Therefore, it is more reasonable to use both $\overrightarrow{\bm{E}}$ and $\overleftarrow{\bm{E}}$ in order to encode the equal information among nodes.

\subsection{Numerical Reasoning}
\label{sec:reasoning}
As we built the graph $\mathcal{G}=(\bm{V},\bm{E})$, we leverage NumGNN to perform reasoning, which is corresponding to the function $\texttt{Reasoning}(\cdot)$ in Eq.~\ref{eqn:reasoing}. The reasoning process is as follows:

\paragraph{Initialization} For each node $v^P_i\in \bm{V}^P$, its representation is initialized as the corresponding column vector of $\bm{M}^P$. Formally, the initial representation is $\bm{v}_i^P=\bm{M}^P[I^P(v_i^P)]$, where $I^P(v^P_i)$ denotes the word index corresponding to $v_i^P$. Similarly, the initial representation $\bm{v}_j^Q$ for a node $v^Q_j\in \bm{V}^Q$ is set as the corresponding column vector of $\bm{M}^Q$. We denote all the initial node representations as $\bm{v}^0=\{\bm{v}_i^P\}\cup\{\bm{v}_j^Q\}$.

\paragraph{One-step Reasoning} 
Given the graph $\mathcal{G}$ and the node representations $\bm{v}$, we use a GNN to perform reasoning in three steps:

(1) {\bf Node Relatedness Measure}: As only a few numbers are relevant for answering a question generally, we compute a weight for each node to by-pass irrelevant numbers in reasoning. Formally, the weight for node $v_i$ is computed as:
    \begin{equation}
        \alpha_i=\mathrm{sigmoid}(\bm{W}_v\bm{v}[i] + b_v),
        \label{eqn:reasoning-begin}
    \end{equation}
where $\bm{W}_v$ is a weight matrix, and $b_v$ is a bias.

(2) {\bf Message Propagation}: As the role a number plays in reasoning is not only decided by itself, but also related to the context, we propagate messages from each node to its neighbors to help to perform reasoning.
As numbers in question and passage may play different roles in reasoning and edges corresponding to different numerical relations should be distinguished, we use relation-specific transform matrices in the message propagation.
Formally, we define the following propagation function for calculating the forward-pass update of a node:
    \begin{equation}
        \widetilde{\bm{v}}'_i = \frac{1}{|\mathcal{N}_i|}\left(\sum_{j \in \mathcal{N}_i }\alpha_j\bm{W}^{\texttt{r}_{ji}}\bm{v}[j]\right),
    \end{equation}
where $\widetilde{\bm{v}}'_i$ is the message representation of node $v_i$, $\texttt{r}_{ji}$ is the relation assigned to edge $e_{ji}$, $\bm{W}^{\texttt{r}_{ji}}$ are relation-specific transform matrices, and $\mathcal{N}_i=\{j|(v_j,v_i)\in\bm{E}\}$ is the neighbors of node $v_i$. 

For each edge $e_{ji}$, $\texttt{r}_{ji}$ is determined by the following two attributes:
\begin{itemize}
    \item {Number relation}: $>$ or $\leq$; 
    \item {Node types}: the two nodes of the edge corresponding to two numbers that: (1) both from the question ($\text{q-q}$); (2) both from the passage ($\text{p-p}$); (3) from the question and the passage respectively  ($\text{q-p}$); (4) from the passage and the question respectively  ($\text{p-q}$).
\end{itemize}
Formally, $\texttt{r}_{ij}\in\{>,\leq\}\times\{\text{q-q},\text{p-p},\text{q-p},\text{p-q}\}$.

(3) {\bf Node Representation Update}: As the message representation obtained in the previous step only contains information from the neighbors, it needs to be fused with the node representation to combine with the information carried by the node itself, which is performed as:
    \begin{equation}
        \bm{v}'_i = \mathrm{ReLU}(\bm{W}_f\bm{v}_i + \widetilde{\bm{v}}'_i + \bm{b}_f),
        \label{eqn:reasoning-end}
    \end{equation}
where $\bm{W}_f$ is a weight matrix, and $\bm{b}_f$ is a bias vector.

We denote the entire one-step reasoning process (Eq.~\ref{eqn:reasoning-begin}-\ref{eqn:reasoning-end}) as a single function 
\begin{equation}
    \bm{v}'=\texttt{Reasoning-Step}(\mathcal{G}, \bm{v}).
\end{equation}

As the graph $\mathcal{G}$ constructed in Sec.~\ref{sec:graph} has encoded the numerical relations via its topology, the reasoning process is numerically-aware.

\paragraph{Multi-step Reasoning}  By single-step reasoning, we can only infer relations between adjacent nodes. However, relations between multiple nodes may be required for certain tasks, e.g., sorting. Therefore, it is essential to perform multi-step reasoning, which can be done as follows：
\begin{equation}
    \bm{v}^t=\texttt{Reasoning-Step}(\bm{v}^{t-1}),
\end{equation}
where $t\ge 1$.
Suppose we perform $K$ steps of reasoning, $\bm{v}^K$ is used as $\bm{U}$ in Eq.~\ref{eqn:reasoing}.

\section{Experiments}
\label{sec:exp}

\subsection{Dataset and Evaluation Metrics}

We evaluate our proposed model on DROP dataset~\citep{dua2019drop}, which is a public numerical MRC dataset. The DROP dataset is constructed by crowd-sourcing, which asks the annotators to generate question-answer pairs according to the given Wikipedia passages, which require numerical reasoning such as addition, counting, or sorting over numbers in the passages. 
There are $77,409$ training samples, $9,536$ development samples and $9,622$ testing samples in the dataset.

In this paper, we adopt two metrics including Exact Match (EM) and numerically-focused F1 scores to evaluate our model following \citet{dua2019drop}. The numerically-focused F1 is set to be $0$ when the predicted answer is mismatched for those questions with the numeric golden answer.

\subsection{Baselines}
For comparison, we select several public models as baselines including \textbf{semantic parsing models}:

\begin{itemize}[topsep=2pt, itemsep=0pt]

\item {Syn Dep}~\citep{dua2019drop}, the neural semantic parsing model (KDG)~\citep{krishnamurthy2017kdg} with Stanford dependencies based sentence representations;

\item {OpenIE}~\citep{dua2019drop}, KDG with open information extraction based sentence representations;

\item {SRL}~\citep{dua2019drop}, KDG with semantic role labeling based sentence representations;
\end{itemize}
\noindent and \textbf{traditional MRC models}:
\begin{itemize}[topsep=2pt, itemsep=0pt]
\item {BiDAF}~\citep{seo2017bidirectional}, an MRC model which utilizes a bi-directional attention flow network to encode the question and passage;

\item {QANet}~\citep{yu2018qanet},  which utilizes convolutions and self-attentions as the building blocks of encoders to represent the question and passage;

\item {BERT}~\citep{devlin2019bert}, a pre-trained bidirectional Transformer-based language model which achieves state-of-the-art performance on lots of public MRC datasets recently;
\end{itemize}
\noindent and \textbf{numerical MRC models}:
\begin{itemize}[topsep=2pt, itemsep=0pt]
\item {NAQANet}~\citep{dua2019drop}, a numerical version of QANet model.

\item {NAQANet+}, an enhanced version of NAQANet implemented by ourselves, which further considers real number (e.g. ``2.5''), richer arithmetic expression, data augmentation, etc. The enhancements are also used in our \OurModel model and the details are given in the Appendix.
\end{itemize}

\subsection{Experimental Settings}

In this paper, we tune our model on the development set and use a grid search to determine the optimal parameters. The dimensions of all the representations (e.g., $\bm{Q}$, $\bm{P}$, $\bm{M}^Q$, $\bm{M}^P$, $\bm{U}$, $\bm{M}_0'$, $\bm{M}_0$ and $\bm{v}$) are set to $128$. If not specified, the reasoning step $K$ is set to $3$. Since other parameters have little effect on the results, we simply follow the settings used in~\citep{dua2019drop}.

We use the Adam optimizer~\citep{kingma2014adam} with $\beta_1=0.8$, $\beta_2=0.999$, $\epsilon=10^{-7}$ to minimize the objective function. The learning rate is $5 \times 10^{-4}$, L2 weight decay $\lambda$ is $10^{-7}$ and the maximum norm value of gradient clipping is $5$. We also apply exponential moving average with a decay rate $0.9999$ on all trainable variables. The model is trained with a batch size of $16$ for $40$ epochs. Passages and questions are trimmed to $400$ and $50$ tokens respectively during training, and trimmed to $1,000$ and $100$ tokens respectively during prediction \footnote{Trimming passages/questions introduces little impact because it only affects about 8\% of the samples.}.

\subsection{Overall Results}

\begin{table}
  \centering
  \small
  \setlength{\tabcolsep}{5pt}
  \begin{tabular}{lcccc}
    \toprule
    \multicolumn{1}{c}{\multirow{2}{*}{Method}}  & \multicolumn{2}{c}{Dev}        &\multicolumn{2}{c}{Test}\\
    \cmidrule(r){2-3}  \cmidrule(r){4-5}
                & EM        & F1        & EM        & F1\\
    \midrule
    \bf{Semantic Parsing}\\
    $\quad$Syn Dep     & $\ $9.38      & 11.64     & $\ $8.51      & 10.84\\
    $\quad$OpenIE      & $\ $8.80      & 11.31     & $\ $8.53      & 10.77\\
    $\quad$SRL         & $\ $9.28      & 11.72     & $\ $8.98      & 11.45\\
    \midrule
    \bf{Traditional MRC}\\
    $\quad$BiDAF       & 26.06     & 28.85     & 24.75     & 27.49 \\
    $\quad$QANet       & 27.50     & 30.44     & 25.50     & 28.36 \\
    $\quad$BERT        & 30.10     & 33.36     & 29.45     & 32.70 \\
    \midrule
    \bf{Numerical MRC}\\
    $\quad$NAQANet     & 46.20     & 49.24     & 44.07     & 47.01 \\
    $\quad$NAQANet+    & 61.47     & 64.85     & 60.82     & 64.29     \\ 
    $\quad$\bf{\OurModel}   & \bf{64.92}    & \bf{68.31}     & \bf{64.56}     & \bf{67.97} \\
    \midrule
    \textbf{Human Performance}
                & -         & -     & 94.09     & 96.42\\
    \bottomrule
  \end{tabular}
  \caption{Overall results on the development and test set. The evaluation metrics are calculated as the maximum over a golden answer set. All the results except ``NAQANet+'' and ``\OurModel{}'' are obtained from~\cite{dua2019drop}.}
  \label{tab:overall-results}
\end{table}

The performance of our \OurModel model and other baselines on DROP dataset are shown in Table~\ref{tab:overall-results}. From the results, we can observe that:

(1) Our \OurModel model achieves better results on both the development and testing sets on DROP dataset as compared to semantic parsing-based models, traditional MRC models and even numerical MRC models NAQANet and NAQANet+. The reason is that our \OurModel model can make full use of the numerical comparison information over numbers in both question and passage via the proposed NumGNN module.

(2) Our implemented NAQANet+ has a much better performance compared to the original version of NAQANet. It verifies the effectiveness of our proposed enhancements for baseline. 

\subsection{Effect of GNN Structure}
\label{sec:exp:structure}
In this part, we investigate the effect of different GNN structures on the DROP development set. The results are shown in Table~\ref{tab:gnn-structure}. The ``Comparison'', ``Number'' and ``ALL'' are corresponding to the comparing question subset
\footnote{We find that many comparing questions in the DROP dataset are biased, of which the answers are the former candidates in the questions. Hence, we employ crowdsourced workers to identify and rewrite all comparing questions to construct an enhanced development set. Specially, for those comparing questions containing answer candidates, we also ask the crowdsourced workers to swap the candidates manually to enlarge the dataset.},
the number-type answer subset, and the entire development set, respectively \footnote{Note that the ``ALL'' result is not the average of ``Comparison'' and ``Number''. It is the performance on the entire development set which also includes questions of selection type, coreference resolution type, etc.}. If we replace the proposed numerically-aware graph (Sec.~\ref{sec:graph}) with a fully connected graph, our model fallbacks to a traditional GNN, denoted as ``GNN'' in the table. Moreover, ``- question num'' denotes the numbers in the question is not included in the graph, and ``- $\leq$ type edge'' and ``- $>$ type edge'' denote edges of $\leq$ and $>$ types are not adopted respectively.

As shown in Table~\ref{tab:gnn-structure}, our proposed NumGNN leads to statistically significant improvements compared to traditional GNN on both EM and F1 scores especially for comparing questions. It indicates that   considering the comparing information over numbers could effectively help the numerical reasoning for comparing questions. Moreover, we find that the numbers in the question are often related to the numerical reasoning for answering the question, thus considering numbers in questions in NumGNN achieves better performance. And the results also justify that encoding ``greater relation'' and ``lower or equal relation'' simultaneously in the graph also benefits our model.

\begin{table}
  \centering
  \small
  \setlength{\tabcolsep}{2.5pt}
  \begin{tabular}{lcccccc}
    \toprule
    \multicolumn{1}{c}{\multirow{2}{*}{Method}}  & \multicolumn{2}{c}{Comparison}        &\multicolumn{2}{c}{Number} & \multicolumn{2}{c}{ALL} \\
    \cmidrule(r){2-3}  \cmidrule(r){4-5} \cmidrule(r){6-7}
                                        & EM    & F1    & EM    & F1    & EM    & F1\\
    \midrule
    GNN                                 & 69.86 & 75.91 & 67.77 & 67.78 & 61.90 & 65.16\\
    NumGNN                              & 74.53 & 80.36 & 69.74 & 69.75 & 64.54 & 68.02\\
    $\quad$- question num               & 74.84 & 80.24 & 68.42 & 68.43 & 63.78 & 67.17\\
    $\quad$- $\leq$ type edge           & 74.89 & 80.51 & 68.48 & 68.50 & 63.66 & 67.06\\
    $\quad$- $>$ type edge              & 74.86 & 80.19 & 68.77 & 68.78 & 63.64 & 66.96\\

    \bottomrule
  \end{tabular}
  \caption{Performance with different GNN structure. ``Comparison'', ``Number'' and ``ALL'' denote the comparing question subset, the number-type answer subset, and the entire development set, respectively.}
  \label{tab:gnn-structure}
\end{table}

\begin{figure}
    \centering
    \includegraphics[width=0.9\columnwidth]{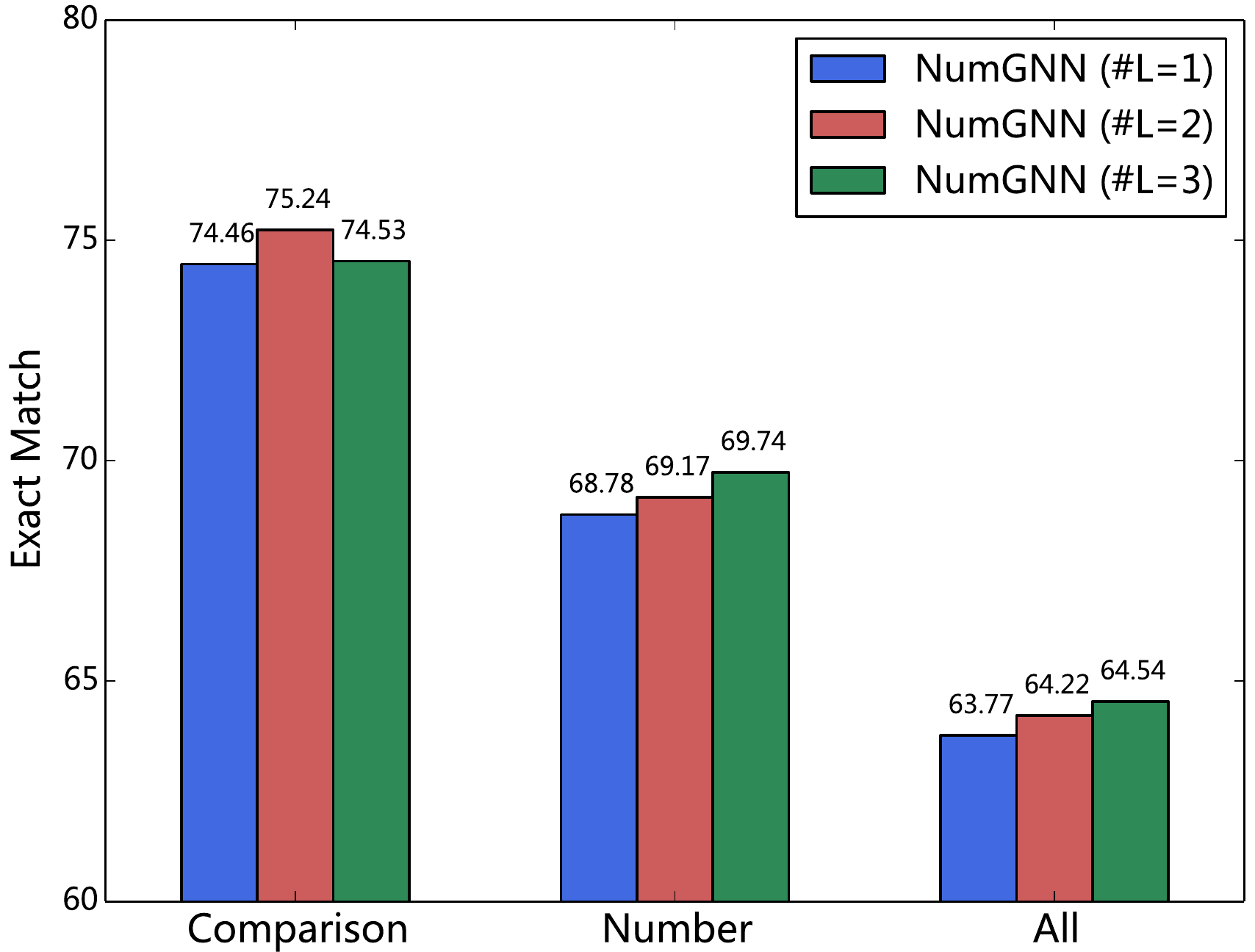}
    \caption{Effect of GNN layer numbers (\# L).}
    \label{fig:layer-number}
\end{figure}

\begin{table*}
	\small
	\center
	\scalebox{0.97}{
	\begin{tabular}{p{0.19\textwidth}p{0.54\textwidth}p{0.08\textwidth}c}
		\toprule
		\textbf{Question \& Answer} & \textbf{Passage} & \textbf{NAQANet+} & \textbf{\OurModel}\\ 
		\midrule
		\multirow{3}{0.18\textwidth}{{\bf Q:} Which age group is larger: under the age of 18 or 18 and 24?}
		& \multirow{5}{0.54\textwidth}{The median age in the city was 22.1 years. \textit{\color{red}10.1\%} of residents were under the age of 18; \textit{\color{red}56.2\%} were between the ages of 18 and 24; 16.1\% were from 25 to 44; 10.5\% were from 45 to 64; and 7\% were 65 years of age or older. The gender makeup of the city was 64.3\% male and 35.7\% female.} & under the age of 18 & 18 and 24\\
		\\\\
		{\bf A:} 18 and 24\\
		\midrule
		\multirow{4}{0.19\textwidth}{{\bf Q}: How many more yards was Longwell's longest field goal over his second longest one?}
		& \multirow{6}{0.54\textwidth}{... The Vikings would draw first blood with a \textit{\color{red}26-yard field goal} by kicker Ryan Longwell. In the second quarter, Carolina got a field goal with opposing kicker John Kasay. The Vikings would respond with another Longwell field goal (\textit{\color{red}a 22-yard FG}) ... In OT, Longwell booted the game-winning \textit{\color{red}19-yard field goal} to give Minnesota the win. It was the first time in Vikings history that a coach ...}
	   & 26-19 = 7   & 26-22 = 4\\
	   \\\\\\\\
		{\bf A:} 26-22=4\\
		\bottomrule
		
	\end{tabular}}
	\caption{Cases from the DROP dataset. We demonstrate the predictions of NAQANet+ and our \OurModel model. Note that the two models only output the arithmetic expressions but we also provide their results for clarity.}
	\label{tab:case}
\end{table*}

\begin{table*}
	\small
	\center
	\scalebox{0.97}{
	\begin{tabular}{p{0.19\textwidth} p{0.55\textwidth}p{0.06\textwidth}c}
		\toprule
		\textbf{Question}  &\textbf{Passage} & \textbf{Answer} & \textbf{\OurModel}\\ 
		\midrule
		Which ancestral groups are at least 10\%?  & As of the census of 2000, there were 7,791 people, 3,155 households, and 2,240 families residing in the county. ... 33.7\% were of \textit{\color{red}Germans}, 13.9\% \textit{\color{red}Swedish} people, 10.1\% \textit{\color{red}Irish} people, 8.8\% United States, 7.0\% English people and 5.4\% Danish people ancestry ... & German; Swedish; Irish & Irish\\
		\midrule
		Were more people 40 and older or 19 and younger?  & Of Saratoga Countys population in 2010, \textit{\color{blue}6.3\%} were between ages of 5 and 9 years, \textit{\color{blue}6.7\%} between 10 and 14 years, \textit{\color{blue}6.5\%} between 15 and 19 years, ... , \textit{\color{magenta}7.9\%} between 40 and 44 years, \textit{\color{magenta}8.5\%} between 45 and 49 years, \textit{\color{magenta}8.0\%} between 50 and 54 years, \textit{\color{magenta}7.0\%} between 55 and 59 years, \textit{\color{magenta}6.4\%} between 60 and 64 years, and \textit{\color{magenta}13.7\%} of age 65 years and over ... & 40 and older & 19 and younger\\
		\bottomrule
	\end{tabular}}
	\caption{Typical error examples. Row 1: the answer is multiple nonadjacent spans; Row 2: Intermediate numbers are involved in reasoning.}
	\label{tab:error}
\end{table*}

\subsection{Effect of GNN Layer Number}
The number of NumGNN layers represents the numerical reasoning ability of our models. A $K$-layer version has the ability for $K$-step numerical inference. In this part, we additionally perform experiments to understand the values of the numbers of NumGNN layers. From Figure \ref{fig:layer-number}, we could observe that:

(1) The $2$-layer version of \OurModel achieves the best performance for the comparing questions. From careful analysis, we find that most comparing questions only require at most $2$-step reasoning (e.g., ``\emph{Who was the second oldest player in the MLB, Clemens or Franco?}''), and therefore the $3$-layer version of \OurModel is more complex but brings no gains for these questions.

(2) The performance of our \OurModel model on the overall development set is improved consistently as the number of GNN layers increases. The reason is that some of the numerical questions require reasoning over many numbers in the passage, which could benefit from the multi-step reasoning ability of multi-layer GNN. However, further investigation shows that the performance gain is not stable when $K\geq 4$. We believe it is due to the intrinsic over smoothing problem of GNNs~\citep{li2018gcnsmoothing}.

\subsection{Case Study}

We further give some examples to show why incorporating comparing information over numbers in the passage could help numerical reasoning in MRC in Table \ref{tab:case}. For the first case, we observe that NAQANet+ gives a wrong prediction, and we find that NAQANet+ will give the same prediction for the question ``\emph{Which age group is \underline{smaller}: under the age of 18 or 18 and 24?}''.  The reason is that NAQANet+ cannot distinguish which one is larger for $10.1\%$ and $56.2\%$. For the second case, NAQANet+ cannot recognize the second longest field goal is $22$-yard and also gives a wrong prediction. For these two cases, our \OurModel model could give the correct answer through the numeric reasoning, which indicates the effectiveness of our \OurModel model.

\subsection{Error Analysis}

To investigate how well our \OurModel model handles sorting/comparison questions and better understand the remaining challenges, we perform an error analysis on a random sample of \OurModel predictions. We find that:

(1) Our \OurModel model can answer about 76\% of sorting/comparison questions correctly, which indicates that our \OurModel model has achieved numerical reasoning ability to some extend.

(2) Among the incorrectly answered sorting/comparison questions, the most ones (26\%) are those whose golden answers are multiple nonadjacent spans (row 1 in Table~\ref{tab:error}), and the second most ones (19\%) are those involving comparison with an intermediate number that does not literally occur in the document/question but has to be derived from counting or arithmetic operation (row 1 in Table~\ref{tab:error}).

\subsection{Discussion}
\label{sec:discussion}
By combining the numerically-aware graph and the NumGNN together, our \OurModel model achieves the numerical reasoning ability. On one hand, the numerically-aware graph encodes numbers as nodes and relationships between them as the edges, which is required for numerical comparison. On the other hand, through one-step reasoning, our NumGNN could perform comparison and identify the numerical condition. After multiple-step reasoning, our NumGNN could further perform sorting.

However, since the numerically-aware graph is pre-defined, our \OurModel is not applicable to the case where an intermediate number has to be derived (e.g., from arithmetic operation) in the reasoning process, which is a major limitation of our model. 

\section{Conclusion and Future Work}
\label{sec:conclusion}
Numerical reasoning skills such as addition, subtraction, sorting and counting are naturally required by machine reading comprehension (MRC) problems in practice. Nevertheless, these skills are not taken into account explicitly for most existing MRC models. In this work, we propose a numerical MRC model named \OurModel which performs explicit numerical reasoning while reading the passages. To be specific, \OurModel encodes the numerical relations among numbers in the question and passage into a graph as its topology, and leverages a numerically-aware graph neural network to perform numerical reasoning on the graph. Our \OurModel model outperforms strong baselines with a large margin on the DROP dataset.

In the future, we will explore the following directions: (1)As we use a pre-defined reasoning graph in our model, it is incapable of handling reasoning process which involves intermediate numbers that not presented in the graph. How to incorporate dynamic graph into our model is an interesting problem. (2) Compared with methods proposed for arithmetic word problems (AWPs), our model has better natural language understanding ability. However, the methods for AWPs  can handle much richer arithmetic expressions. Therefore, how to combine both of their abilities to develop a more powerful numerical MRC model is an interesting future direction. (3) Symbolic reasoning plays a crucial role in human reading comprehension. Our work integrates numerical reasoning, which is a special case of symbolic reasoning, into traditional MRC systems. How to incorporate more sophisticated symbolic reasoning abilities into MRC systems is also a valuable future direction.

\section*{Acknowledgments}
We would like to thank all anonymous reviewers for their insightful comments, and thank Yan Zhang for her help on improving the presentation of Figure~\ref{fig:main}.

\begin{table*}[h!]
	\centering
	\small
	
	\begin{tabular}{lcccccc}
		\toprule
		\multicolumn{1}{c}{\multirow{2}{*}{Method}}  & \multicolumn{2}{c}{Comparison}        &\multicolumn{2}{c}{Number} & \multicolumn{2}{c}{ALL} \\
		\cmidrule(r){2-3}  \cmidrule(r){4-5} \cmidrule(r){6-7}
		& EM    & F1    & EM    & F1    & EM    & F1\\
		\midrule
		NAQANet+                               & 69.11 & 75.62 & 66.92 & 66.94 & 61.11 & 64.54\\
		$\quad$ - real number                  & 66.87 & 73.25 & 45.82 & 45.85 & 47.82 & 51.22\\
		$\quad$ - richer arithmetic expression & 68.62 & 74.55 & 52.48 & 52.51 & 52.02 & 55.32\\
		
		$\quad$ - passage-preferred            & 64.06 & 72.34 & 66.46 & 66.47 & 59.64 & 63.34\\
		$\quad$ - data augmentation            & 65.28 & 71.81 & 67.05 & 67.07 & 61.21 & 64.60\\
		\bottomrule
	\end{tabular}
	\caption{Baseline enhancements ablation.}
	\label{tab:tricks}
\end{table*}

\section*{Appendix: Baseline Enhancements}
\label{sec:tricks}

The major enhancements leveraged by our implemented NAQANet+ model
include:

(1) ``real number'': Unlike NAQANet only considers integer numbers, we also consider real numbers.

(2) ``richer arithmetic expression'': We conceptually append an extra number ``100'' to the passage to support arithmetic expressions like ``100-25'', which is required for answering questions such as ``\emph{How many percent were not American?}''.

(3) ``passage-preferred'': If an answer is both a span of the question and the passage, we only propagate gradients through the output layer for processing ``Passage span'' type answers.

(4) ``data augmentation'': The original questions in the DROP dataset are generated by crowdsourced workers. For the comparing questions which contain answer candidates, we observe that the workers frequently only change the incorrect answer candidate to generate a new question. For example, ``\emph{How many from the census is bigger: \underline{Germans} or \underline{English}?}'' whose golden answer is ``Germans'' is modified to ``\emph{How many from the census is bigger: \underline{Germans} or \underline{Irish}?}''. This may introduce undesired inductive bias to the model. Therefore, we propose to augment the training dataset with new questions automatically generated by swapping the candidate answers, e.g., ``\emph{How many from the census is bigger: \underline{English} or \underline{Germans}?}'' is added to the training dataset.

We further conduct ablation studies on the enhancements. And the validation scores on the development set are shown in Table~\ref{tab:tricks}.
As can be seen from Table~\ref{tab:tricks}:

(1) The uses of real number and richer arithmetic expression are crucial for answering numerical questions: both EM and F1 drop drastically by up to $15-21$ points if they are removed. 

(2) The passage-preferred strategy and data augmentation are also necessary components that contribute significant improvements for those comparing questions.

\bibliography{emnlp-ijcnlp-2019}

\begin{thebibliography}{26}
\expandafter\ifx\csname natexlab\endcsname\relax\def\natexlab#1{#1}\fi

\bibitem[{Chen et~al.(2016)Chen, Bolton, and
  Manning}]{chen-bolton-manning:2016:P16-1}
Danqi Chen, Jason Bolton, and Christopher~D. Manning. 2016.
\newblock \href {https://doi.org/10.18653/v1/P16-1223} {A thorough examination
  of the {CNN}/{Daily} {Mail} reading comprehension task}.
\newblock In \emph{Proceedings of the 54th Annual Meeting of the Association
  for Computational Linguistics (Volume 1: Long Papers)}, pages 2358--2367,
  Berlin, Germany.

\bibitem[{Cui et~al.(2017)Cui, Chen, Wei, Wang, Liu, and
  Hu}]{cui-EtAl:2017:Long}
Yiming Cui, Zhipeng Chen, Si~Wei, Shijin Wang, Ting Liu, and Guoping Hu. 2017.
\newblock \href {https://doi.org/10.18653/v1/P17-1055}
  {Attention-over-attention neural networks for reading comprehension}.
\newblock In \emph{Proceedings of the 55th Annual Meeting of the Association
  for Computational Linguistics (Volume 1: Long Papers)}, pages 593--602,
  Vancouver, Canada.

\bibitem[{Devlin et~al.(2019)Devlin, Chang, Lee, and
  Toutanova}]{devlin2019bert}
Jacob Devlin, Ming-Wei Chang, Kenton Lee, and Kristina Toutanova. 2019.
\newblock \href {https://doi.org/10.18653/v1/N19-1423} {{BERT}: Pre-training of
  deep bidirectional transformers for language understanding}.
\newblock In \emph{Proceedings of the 2019 Conference of the North {A}merican
  Chapter of the Association for Computational Linguistics: Human Language
  Technologies, Volume 1 (Long and Short Papers)}, pages 4171--4186,
  Minneapolis, Minnesota.

\bibitem[{Dhingra et~al.(2017)Dhingra, Liu, Yang, Cohen, and
  Salakhutdinov}]{dhingra-EtAl:2017:Long2}
Bhuwan Dhingra, Hanxiao Liu, Zhilin Yang, William Cohen, and Ruslan
  Salakhutdinov. 2017.
\newblock \href {https://doi.org/10.18653/v1/P17-1168} {Gated-attention readers
  for text comprehension}.
\newblock In \emph{Proceedings of the 55th Annual Meeting of the Association
  for Computational Linguistics (Volume 1: Long Papers)}, pages 1832--1846,
  Vancouver, Canada.

\bibitem[{Dua et~al.(2019)Dua, Wang, Dasigi, Stanovsky, Singh, and
  Gardner}]{dua2019drop}
Dheeru Dua, Yizhong Wang, Pradeep Dasigi, Gabriel Stanovsky, Sameer Singh, and
  Matt Gardner. 2019.
\newblock \href {https://doi.org/10.18653/v1/N19-1246} {{DROP}: A reading
  comprehension benchmark requiring discrete reasoning over paragraphs}.
\newblock In \emph{Proceedings of the 2019 Conference of the North {A}merican
  Chapter of the Association for Computational Linguistics: Human Language
  Technologies, Volume 1 (Long and Short Papers)}, pages 2368--2378,
  Minneapolis, Minnesota.

\bibitem[{Hermann et~al.(2015)Hermann, Kocisky, Grefenstette, Espeholt, Kay,
  Suleyman, and Blunsom}]{hermann2015teaching}
Karl~Moritz Hermann, Tomas Kocisky, Edward Grefenstette, Lasse Espeholt, Will
  Kay, Mustafa Suleyman, and Phil Blunsom. 2015.
\newblock \href
  {http://papers.nips.cc/paper/5945-teaching-machines-to-read-and-comprehend.pdf}
  {Teaching machines to read and comprehend}.
\newblock In \emph{Proceedings of Advances in Neural Information Processing
  Systems}, pages 1693--1701.

\bibitem[{Hosseini et~al.(2014)Hosseini, Hajishirzi, Etzioni, and
  Kushman}]{hosseini2014learning}
Mohammad~Javad Hosseini, Hannaneh Hajishirzi, Oren Etzioni, and Nate Kushman.
  2014.
\newblock \href {https://doi.org/10.3115/v1/D14-1058} {Learning to solve
  arithmetic word problems with verb categorization}.
\newblock In \emph{Proceedings of the 2014 Conference on Empirical Methods in
  Natural Language Processing ({EMNLP})}, pages 523--533, Doha, Qatar.

\bibitem[{Huang et~al.(2016)Huang, Shi, Lin, Yin, and Ma}]{huang2016well}
Danqing Huang, Shuming Shi, Chin-Yew Lin, Jian Yin, and Wei-Ying Ma. 2016.
\newblock \href {https://doi.org/10.18653/v1/P16-1084} {How well do computers
  solve math word problems? {Large}-scale dataset construction and evaluation}.
\newblock In \emph{Proceedings of the 54th Annual Meeting of the Association
  for Computational Linguistics (Volume 1: Long Papers)}, pages 887--896,
  Berlin, Germany.

\bibitem[{Joshi et~al.(2017)Joshi, Choi, Weld, and
  Zettlemoyer}]{joshi-EtAl:2017:Long}
Mandar Joshi, Eunsol Choi, Daniel Weld, and Luke Zettlemoyer. 2017.
\newblock \href {https://doi.org/10.18653/v1/P17-1147} {{T}rivia{QA}: A large
  scale distantly supervised challenge dataset for reading comprehension}.
\newblock In \emph{Proceedings of the 55th Annual Meeting of the Association
  for Computational Linguistics (Volume 1: Long Papers)}, pages 1601--1611,
  Vancouver, Canada.

\bibitem[{Kadlec et~al.(2016)Kadlec, Schmid, Bajgar, and
  Kleindienst}]{kadlec2016text}
Rudolf Kadlec, Martin Schmid, Ond{\v{r}}ej Bajgar, and Jan Kleindienst. 2016.
\newblock \href {https://doi.org/10.18653/v1/P16-1086} {Text understanding with
  the attention sum reader network}.
\newblock In \emph{Proceedings of the 54th Annual Meeting of the Association
  for Computational Linguistics (Volume 1: Long Papers)}, pages 908--918,
  Berlin, Germany.

\bibitem[{Kaushik and Lipton(2018)}]{kaushik2018much}
Divyansh Kaushik and Zachary~C. Lipton. 2018.
\newblock \href {https://www.aclweb.org/anthology/D18-1546} {How much reading
  does reading comprehension require? {A} critical investigation of popular
  benchmarks}.
\newblock In \emph{Proceedings of the 2018 Conference on Empirical Methods in
  Natural Language Processing}, pages 5010--5015, Brussels, Belgium.

\bibitem[{Kingma and Ba(2015)}]{kingma2014adam}
Diederik~P Kingma and Jimmy Ba. 2015.
\newblock Adam: A method for stochastic optimization.
\newblock In \emph{Proceedings of ICLR 2015 : International Conference on
  Learning Representations 2015}.

\bibitem[{Koncel-Kedziorski et~al.(2015)Koncel-Kedziorski, Hajishirzi,
  Sabharwal, Etzioni, and Ang}]{koncel2015parsing}
Rik Koncel-Kedziorski, Hannaneh Hajishirzi, Ashish Sabharwal, Oren Etzioni, and
  Siena~Dumas Ang. 2015.
\newblock \href {https://doi.org/10.1162/tacl_a_00160} {Parsing algebraic word
  problems into equations}.
\newblock \emph{Transactions of the Association for Computational Linguistics},
  3:585--597.

\bibitem[{Krishnamurthy et~al.(2017)Krishnamurthy, Dasigi, and
  Gardner}]{krishnamurthy2017kdg}
Jayant Krishnamurthy, Pradeep Dasigi, and Matt Gardner. 2017.
\newblock \href {https://doi.org/10.18653/v1/D17-1160} {Neural semantic parsing
  with type constraints for semi-structured tables}.
\newblock In \emph{Proceedings of the 2017 Conference on Empirical Methods in
  Natural Language Processing}, pages 1516--1526, Copenhagen, Denmark.

\bibitem[{Lai et~al.(2017)Lai, Xie, Liu, Yang, and Hovy}]{lai2017race}
Guokun Lai, Qizhe Xie, Hanxiao Liu, Yiming Yang, and Eduard Hovy. 2017.
\newblock \href {https://doi.org/10.18653/v1/D17-1082} {{RACE}: Large-scale
  {R}e{A}ding comprehension dataset from examinations}.
\newblock In \emph{Proceedings of the 2017 Conference on Empirical Methods in
  Natural Language Processing}, pages 785--794, Copenhagen, Denmark.

\bibitem[{Li et~al.(2018)Li, Han, and Wu}]{li2018gcnsmoothing}
Qimai Li, Zhichao Han, and Xiao-Ming Wu. 2018.
\newblock Deeper insights into graph convolutional networks for semi-supervised
  learning.
\newblock In \emph{Proceedings of the Thirty-Second AAAI Conference on
  Artificial Intelligence}, pages 3538--3545.

\bibitem[{Ling et~al.(2017)Ling, Yogatama, Dyer, and Blunsom}]{wang2017AWP}
Wang Ling, Dani Yogatama, Chris Dyer, and Phil Blunsom. 2017.
\newblock Program induction by rationale generation: Learning to solve and
  explain algebraic word problems.
\newblock \emph{arXiv preprint arXiv:1705.04146}.

\bibitem[{Rajpurkar et~al.(2016)Rajpurkar, Zhang, Lopyrev, and
  Liang}]{rajpurkar2016squad}
Pranav Rajpurkar, Jian Zhang, Konstantin Lopyrev, and Percy Liang. 2016.
\newblock \href {https://doi.org/10.18653/v1/D16-1264} {{SQ}u{AD}: 100,000+
  questions for machine comprehension of text}.
\newblock In \emph{Proceedings of the 2016 Conference on Empirical Methods in
  Natural Language Processing}, pages 2383--2392, Austin, Texas.

\bibitem[{Roy and Roth(2015)}]{roy2015solving}
Subhro Roy and Dan Roth. 2015.
\newblock \href {https://doi.org/10.18653/v1/D15-1202} {Solving general
  arithmetic word problems}.
\newblock In \emph{Proceedings of the 2015 Conference on Empirical Methods in
  Natural Language Processing}, pages 1743--1752, Lisbon, Portugal.

\bibitem[{Seo et~al.(2017)Seo, Kembhavi, Farhadi, and
  Hajishirzi}]{seo2017bidirectional}
Minjoon Seo, Aniruddha Kembhavi, Ali Farhadi, and Hannaneh Hajishirzi. 2017.
\newblock Bidirectional attention flow for machine comprehension.
\newblock In \emph{Proceedings of ICLR 2017: the 5th International Conference
  on Learning Representations}.

\bibitem[{Sugawara et~al.(2018)Sugawara, Inui, Sekine, and
  Aizawa}]{sugawara2018makes}
Saku Sugawara, Kentaro Inui, Satoshi Sekine, and Akiko Aizawa. 2018.
\newblock \href {https://www.aclweb.org/anthology/D18-1453} {What makes reading
  comprehension questions easier?}
\newblock In \emph{Proceedings of the 2018 Conference on Empirical Methods in
  Natural Language Processing}, pages 4208--4219, Brussels, Belgium.

\bibitem[{Wang et~al.(2018)Wang, Zhang, Gao, Song, Guo, and
  Shen}]{wang2018mathdqn}
Lei Wang, Dongxiang Zhang, Lianli Gao, Jingkuan Song, Long Guo, and Heng~Tao
  Shen. 2018.
\newblock {MathDQN}: Solving arithmetic word problems via deep reinforcement
  learning.
\newblock In \emph{Proceedings of the Thirty-Second AAAI Conference on
  Artificial Intelligence (AAAI-18)}.

\bibitem[{Wang et~al.(2017{\natexlab{a}})Wang, Yang, Wei, Chang, and
  Zhou}]{wang2017gated}
Wenhui Wang, Nan Yang, Furu Wei, Baobao Chang, and Ming Zhou.
  2017{\natexlab{a}}.
\newblock \href {https://doi.org/10.18653/v1/P17-1018} {Gated self-matching
  networks for reading comprehension and question answering}.
\newblock In \emph{Proceedings of the 55th Annual Meeting of the Association
  for Computational Linguistics (Volume 1: Long Papers)}, pages 189--198,
  Vancouver, Canada.

\bibitem[{Wang et~al.(2017{\natexlab{b}})Wang, Liu, and
  Shi}]{wang-etal-2017-deep-neural}
Yan Wang, Xiaojiang Liu, and Shuming Shi. 2017{\natexlab{b}}.
\newblock \href {https://doi.org/10.18653/v1/D17-1088} {Deep neural solver for
  math word problems}.
\newblock In \emph{Proceedings of the 2017 Conference on Empirical Methods in
  Natural Language Processing}, pages 845--854, Copenhagen, Denmark.

\bibitem[{Xiong et~al.(2017)Xiong, Zhong, and Socher}]{xiong2017dynamic}
Caiming Xiong, Victor Zhong, and Richard Socher. 2017.
\newblock Dynamic coattention networks for question answering.
\newblock In \emph{Proceedings of ICLR 2017: the 5th International Conference
  on Learning Representations}.

\bibitem[{Yu et~al.(2018)Yu, Dohan, Luong, Zhao, Chen, Norouzi, and
  Le}]{yu2018qanet}
Adams~Wei Yu, David Dohan, Minh-Thang Luong, Rui Zhao, Kai Chen, Mohammad
  Norouzi, and Quoc~V Le. 2018.
\newblock {QANet}: Combining local convolution with global self-attention for
  reading comprehension.
\newblock In \emph{Proceedings of ICLR 2018: the 6th International Conference
  on Learning Representations}.

\end{thebibliography}
\bibliographystyle{acl_natbib}

\end{document}


\maketitle

\appendix
\begin{table*}[h!]
  \centering
  \small
  
  \begin{tabular}{lcccccc}
    \toprule
    \multicolumn{1}{c}{\multirow{2}{*}{Method}}  & \multicolumn{2}{c}{Comparison}        &\multicolumn{2}{c}{Number} & \multicolumn{2}{c}{ALL} \\
    \cmidrule(r){2-3}  \cmidrule(r){4-5} \cmidrule(r){6-7}
                                           & EM    & F1    & EM    & F1    & EM    & F1\\
    \midrule
    NAQANet+                               & 69.11 & 75.62 & 66.92 & 66.94 & 61.11 & 64.54\\
    $\quad$ - real number                  & 66.87 & 73.25 & 45.82 & 45.85 & 47.82 & 51.22\\
    $\quad$ - richer arithmetic expression & 68.62 & 74.55 & 52.48 & 52.51 & 52.02 & 55.32\\
    
    $\quad$ - passage-preferred            & 64.06 & 72.34 & 66.46 & 66.47 & 59.64 & 63.34\\
    $\quad$ - data augmentation            & 65.28 & 71.81 & 67.05 & 67.07 & 61.21 & 64.60\\
    \bottomrule
  \end{tabular}
  \caption{Baseline enhancements ablation.}
  \label{tab:tricks}
\end{table*}

\section{Baseline Enhancements}
\label{sec:tricks}

The major enhancements leveraged by our implemented NAQANet+ model
include:

(1) ``real number'': Unlike NAQANet only considers integer numbers, we also consider real numbers.

(2) ``richer arithmetic expression'': We conceptually append an extra number ``100'' to the passage to support arithmetic expressions like ``100-25'', which is required for answering questions such as ``\emph{How many percent were not American?}''.



(3) ``passage-preferred'': If an answer is both a span of the question and the passage, we only propagate gradients through the output layer for processing ``Passage span'' type answers.

(4) ``data augmentation'': The original questions in the DROP dataset are generated by crowdsourced workers. For the comparing questions which contain answer candidates, we observe that the workers frequently only change the incorrect answer candidate to generate a new question. For example, ``\emph{How many from the census is bigger: \underline{Germans} or \underline{English}?}'' whose golden answer is ``Germans'' is modified to ``\emph{How many from the census is bigger: \underline{Germans} or \underline{Irish}?}''. This may introduce undesired inductive bias to the model. Therefore, we propose to augment the training dataset with new questions automatically generated by swapping the candidate answers, e.g., ``\emph{How many from the census is bigger: \underline{English} or \underline{Germans}?}'' is added to the training dataset.

We further conduct ablation studies on the enhancements. And the validation scores on the development set are shown in Table~\ref{tab:tricks}.
As can be seen from Table~\ref{tab:tricks}:

(1) The uses of real number and richer arithmetic expression are crucial for answering numerical questions: both EM and F1 drop drastically by up to $15-21$ points if they are removed. 

(2) The passage-preferred strategy and data augmentation are also necessary components that contribute significant improvements for those comparing questions.